\documentclass[11pt,a4paper]{article}
\usepackage[hyperref]{acl2020}
\usepackage{times}
\usepackage{latexsym}

\usepackage{graphicx} 
\usepackage{multirow}
\usepackage{amsmath}
\usepackage{amsfonts}

\usepackage{booktabs}
\def\vector#1{\mbox{\boldmath $\mathrm{#1}$}} % \vector
\newcommand{\argmax}{\mathop\textrm{arg~max}\limits} % argmax
\usepackage{microtype}

\aclfinalcopy % Uncomment this line for the final submission
\title{Reflection-based Word Attribute Transfer}
\author{Yoichi Ishibashi, ~~Katsuhito Sudoh, 
~~Koichiro Yoshino, ~~Satoshi Nakamura \\
Nara Institute of Science and Technology \\
\texttt{\{ishibashi.yoichi.ir3, sudoh, koichiro, s-nakamura\}@is.naist.jp} \\
}

\begin{document}
\maketitle
\begin{abstract}
Word embeddings, which often represent such analogic relations as $\overrightarrow{king} - \overrightarrow{man} + \overrightarrow{woman} \approx \overrightarrow{queen}$, 
can be used to change a word's attribute, including its gender.
For transferring \emph{king} into \emph{queen} in this analogy-based manner, 
we subtract a difference vector $\overrightarrow{man} - \overrightarrow{woman}$ based on the knowledge that \emph{king} is male.
However, developing such knowledge is very costly for words and attributes.
In this work, we propose a novel method for word attribute transfer based on reflection mappings without such an analogy operation.
Experimental results show that our proposed method can transfer the word attributes of the given words without changing the words that do not have the target attributes.
\end{abstract}

\section{Introduction}
Word-embedding methods handle word semantics in natural language processing \cite{DBLP:journals/corr/abs-1301-3781, DBLP:conf/nips/MikolovSCCD13, DBLP:conf/emnlp/PenningtonSM14, DBLP:journals/corr/VilnisM14,  DBLP:journals/tacl/BojanowskiGJM17}.
Such word-embedding models as skip-gram with negative sampling \cite[SGNS;][]{DBLP:conf/nips/MikolovSCCD13} or GloVe \cite{DBLP:conf/emnlp/PenningtonSM14} capture such analogic relations
as $\overrightarrow{king} - \overrightarrow{man} + \overrightarrow{woman} \approx \overrightarrow{queen}$. 
Previous work \cite{DBLP:conf/nips/LevyG14, DBLP:journals/tacl/AroraLLMR16, DBLP:conf/acl/GittensAM17, DBLP:conf/acl/EthayarajhDH19a, DBLP:conf/icml/AllenH19} offers theoretical explanation based on Pointwise Mutual Information \cite[PMI;][]{DBLP:journals/coling/ChurchH90} for maintaining analogic relations in word vectors. 

These relations can be used to transfer a certain attribute of a word,
such as changing \emph{king} into \emph{queen} by transferring its gender.
%
%This transfer can be applied to data augmentation; for example, rewriting \emph{He is a boy} to \emph{She is a girl}. 
This transfer can be applied to perform data augmentation; for example, rewriting \emph{He is a boy} to \emph{She is a girl}. 
It can be used to generate negative examples for natural language inference, for example. %%% sudoh; camera-ready
%
%We tackle a novel task that transfers any word of certain attributes: \emph{word attribute transfer}.
We tackle a novel task that transfers any word associated with certain attributes: \emph{word attribute transfer}.

A naive way for word attribute transfer is to use a difference vector based on analogic relations, such as adding $\overrightarrow{woman} - \overrightarrow{man}$ to $\overrightarrow{king}$ to obtain $\overrightarrow{queen}$.
This requires explicit knowledge whether an input word is male or female;
we have to add a difference vector to a male word and subtract it from a female word for the gender transfer.
We also have to avoid changing words without gender attributes,
such as \emph{is} and \emph{a} in the example above,
since they are non-attribute words.
Developing such knowledge is very costly for words and attributes in practice.
In this work, we propose a novel framework for a word attribute transfer based on \emph{reflection} that does not require explicit knowledge of the given words in its prediction.

The contribution of this work is two-fold: (1) We propose a word attribute transfer method that obtains a vector with an inverted binary attribute without explicit knowledge.
%and develop a method based on the definition of an ideal transfer mapping;
%
(2) The proposed method demonstrates more accurate word attribute transfer for words that have target attributes than other baselines without changing the words that do not have the target attributes.
\begin{figure}[t]
   \centering
    \includegraphics[clip, width=7cm]{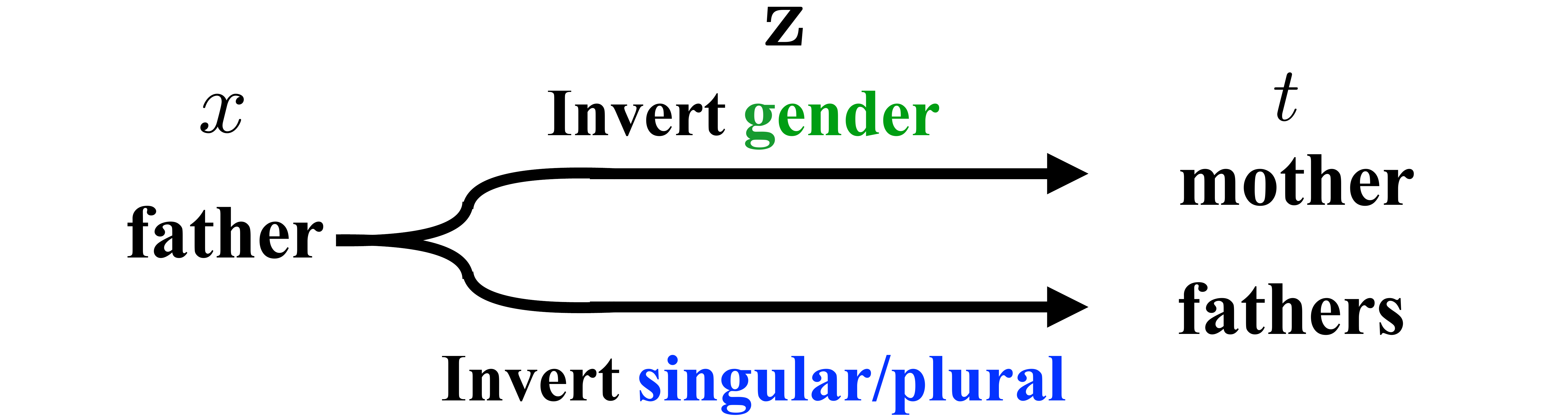}
    \caption{Examples of word attribute transfer}
    \label{fig:concept}
\end{figure}

\section{Word Attribute Transfer Task}
In this task, we focus on modeling the binary attributes (e.g. male and female\footnote[1]{Gender-specific words are sometimes considered socially problematic. Here we use this as an example from the man-woman relation.}). 
%
%This section describes and defines the word attribute transfer task.
%
Let $x$ denote a word and let $\vector{v}_x$ denote its vector representation.
We assume that $\vector{v}_x$ is learned in advance with an embedding model, such as skip-gram.
In this task, we have two inputs, word $x$ and vector $\vector{z}$, which represent a certain target attribute, and output word $t$ with the inverted attribute of $x$ for $\vector{z}$.
In this paper, $\vector{z}$ is a 300-dimensional vector embedded from a target attribute ID using an embedding function of a deep learning framework. 
For example, given a set of attributes $\vector{\mathcal{Z}} = \{\mathrm{gender}, \mathrm{antonym}\}$, we assign different random vectors $\vector{z}_{\mathrm{gender}}$ for gender and $\vector{z}_{\mathrm{antonym}}$ for antonym, respectively.
Let $\mathcal{A}$ denote a set of triplets $(x, t, \vector{z})$, e.g., $(man, woman, \vector{z}_{\mathrm{gender}})\in \mathcal{A}$,
and $\mathcal{N}$ denote a set of words without attribute $\vector{z}$, e.g., $(person, \vector{z}_{\mathrm{gender}}) \in \mathcal{N}$.
This task transfers input word vector $\vector{v}_x$ to target word vector $\vector{v}_t$ by transfer function $f_{\vector{z}}$ that inverts attribute $\vector{z}$ of $\vector{v}_x$:
%
%In other words, output $\vector{v}_y$ should be close to the vector of corresponding target word $\vector{v}_t$, which is typically the nearest neighbor of $\vector{v}_t$ in the word embedding space.
%
\begin{equation}
	\vector{v}_{t} \approx \vector{v}_{y} = f_{\vector{z}} (\vector{v}_x).
	\label{eq:task}
\end{equation}
The following property must be satisfied: (1) attribute words $\{x | (x, t, \vector{z}) \in \mathcal{A}\}$ are transferred to their counterparts and (2) non-attribute words $\{x | (x, \vector{z}) \in \mathcal{N}\}$ are not changed (transferred back into themselves).
For instance with $\vector{z}_{\mathrm{gender}}$, given input word \emph{man}, gender attribute transfer $f_{\vector{z}_{\mathrm{gender}}} (\vector{v}_{man})$ should result in a vector close to $\vector{v}_{woman}$.
Given another input word \emph{person} as $x$, the results should be $\vector{v}_{person}$.

\section{Analogy-based Word Attribute Transfer}
Analogy is a general idea that can be used for word attribute transfer.
%
%Several PMI-based word embedding methods such as SGNS and GloVe \cite{DBLP:conf/conll/LevyG14, DBLP:conf/naacl/MikolovYZ13, DBLP:conf/repeval/Linzen16} aimed to capture analogic relations. %tackled to embed words into word embedding space to capture the analogic relations. 
%
PMI-based word embedding, such as SGNS and GloVe, captures analogic relations, including Eq. \ref{eq:an1} \cite{DBLP:conf/naacl/MikolovYZ13, DBLP:conf/conll/LevyG14, DBLP:conf/repeval/Linzen16}.
%
%For instance, $\vector{v}_{queen}$ is close to the vector obtained on the right side of Eq. \ref{eq:an1}.
%Several PMI-based word embedding methods such as SGNS and GloVe \cite{DBLP:conf/conll/LevyG14, DBLP:conf/naacl/MikolovYZ13, DBLP:conf/repeval/Linzen16} can capture analogic relations such as Eq.\ref{eq:an1}. 
%
By rearranging Eq. \ref{eq:an1}, Eq. \ref{eq:an2} is obtained:
\begin{eqnarray}
	\vector{v}_{queen} & \approx & \vector{v}_{king} - \vector{v}_{man} + \vector{v}_{woman}, \label{eq:an1} \\
	& \approx & \vector{v}_{king} - (\vector{v}_{man} - \vector{v}_{woman} ). \label{eq:an2}
\end{eqnarray}
%
%We can transfer the gender attribute by subtracting difference vector $\vector{v}_{man} - \vector{v}_{woman}$ from input word vectors.%,  e.g., $\vector{v}_{king}$. 
%
The analogy-based transfer function is
\begin{eqnarray}
  f_{\vector{z}}(\vector{v}_x) = 
  \begin{cases}
    \vector{v}_x - \vector{d} & \hspace{2mm} \mathrm{if} \hspace{3mm} x \in \mathcal{M} ,\\
    \vector{v}_x + \vector{d} & \hspace{2mm} \mathrm{if} \hspace{3mm} x \in \mathcal{F},
  \end{cases}
  \label{eq:an3}
\end{eqnarray}
where $\mathcal{M}$ is a set of words with a target attribute (e.g., male) and $\mathcal{F}$ is a set of words with an inverse attribute (e.g., female).
$\vector{d}$ is a difference vector, such as $\vector{v}_{man} - \vector{v}_{woman}$.
Eq. \ref{eq:an3} indicates that the operation changes depending on whether input word $x$ belongs to $\mathcal{M}$ or $\mathcal{F}$.
%
%However, this knowledge cannot be developed for various words and attributes in practice.
%
%However, to transfer the word attribute by analogy, we need such explicit knowledge which an attribute value ($\mathcal{M}$, $\mathcal{F}$ or others) the input word has.
However, to transfer the word attribute by analogy, we need such explicit knowledge as attribute value ($\mathcal{M}$, $\mathcal{F}$ or others) that is contained by the input word.

\section{Reflection-based Word Attribute Transfer}
%
%In chapter 4, we showed that analogy-based word attribute transfer have a problem because the transfer use prior knowledge about input words.
%In this chapter we describe a proposed method that does not use such prior knowledge.
%
\subsection{Ideal Transfer without Knowledge}
What is ideal transfer function $f_{\vector{z}}$ for a word attribute transfer?
The following are the ideal natures of such a transfer function:
\begin{eqnarray}
	\forall (m, w, \vector{z}) \in \mathcal{A}, && \vector{v}_m = f_{\vector{z}}(\vector{v}_w), \label{eq:inv1} \\
	\forall (m, w, \vector{z}) \in \mathcal{A}, && \vector{v}_w = f_{\vector{z}}(\vector{v}_m), \label{eq:inv2} \\
	\forall (u, \vector{z}) \in \mathcal{N}, && \vector{v}_u = f_{\vector{z}}(\vector{v}_u). \label{eq:inv_non} 
\end{eqnarray}
%
%where $m \in \mathcal{M}$ and $w \in \mathcal{F}$.
%
This function $f_{\vector{z}}$ enables a word to be transferred without explicit knowledge
because operation $f_{\vector{z}}$ does not change depending on whether input word belongs to $\mathcal{M}$ or $\mathcal{F}$.
%
%$f_{\vector{z}}$ transfers $\vector{v}_m$ to $\vector{v}_w$ and $\vector{v}_w$ to $\vector{v}_m$ without such explicit knowledge as $m \in \mathcal{M}$ and $w \in \mathcal{F}$.
%
By combining Eqs. \ref{eq:inv1}, \ref{eq:inv2} and \ref{eq:inv_non}, we obtain the following formulas: %カメラレディ：追加2
\begin{eqnarray}
	%\forall m \in \mathcal{M}, && \vector{v}_m = f_{\vector{z}}(\hspace{0.7mm}f_{\vector{z}}(\vector{v}_m)\hspace{0.7mm}), \label{eq:inv3} \\
	%\forall w \in \mathcal{F}, && \vector{v}_w = f_{\vector{z}}(\hspace{0.7mm}f_{\vector{z}}(\vector{v}_w)\hspace{0.7mm}). \label{eq:inv4} 
	\forall (m, w, \vector{z}) \in \mathcal{A}, && \vector{v}_m = f_{\vector{z}}(\hspace{0.7mm}f_{\vector{z}}(\vector{v}_m)\hspace{0.7mm}), \label{eq:inv3} \\
	\forall (m, w, \vector{z}) \in \mathcal{A}, && \vector{v}_w = f_{\vector{z}}(\hspace{0.7mm}f_{\vector{z}}(\vector{v}_w)\hspace{0.7mm}), \label{eq:inv4} \\
	\forall (u, \vector{z}) \in \mathcal{N}, && \vector{v}_u = f_{\vector{z}}(\hspace{0.7mm}f_{\vector{z}}(\vector{v}_u)\hspace{0.7mm}). \label{eq:inv_non2} 
\end{eqnarray}
Hence, the ideal transfer function is a mapping that becomes an identity mapping when we apply it twice for any $\vector{v}$.
Such a mapping is called \textit{involution} in geometry.
For example, $f \colon \vector{v} \mapsto - \vector{v}$ is one example of an involution.
%
%Note that the identity map itself, such as $f \colon \vector{v} \mapsto \vector{v}$, is excluded from the involution.

\subsection{Reflection} 
\textit{Reflection} $\mathrm{Ref}_{\vector{a},\vector{c}}$ is an ideal function because this mapping is an involution:
\begin{eqnarray}
    \forall \vector{v}  \in \mathbb{R}^n, && \vector{v} = \mathrm{Ref}_{\vector{a},\vector{c}}(\hspace{0.7mm} \mathrm{Ref}_{\vector{a},\vector{c}}(\vector{v}) \hspace{0.7mm}).
    \label{eq:doubleRef}
\end{eqnarray}
Reflection reverses the location between two vectors in a Euclidean space through an hyperplane called a \emph{mirror}.
%
%A \textit{reflection} is an involution that reverses the location between two vectors in a Euclidean space through an affine hyperplane (mirror).
%
%Reflection is an ideal function because every point returns to its original location when the reflection is applied twice: 
%
%Reflection is different from an inverse mapping because reflection can transfer both $\vector{v}$ and $\vector{v}'$ with the same reflection mapping.
%
Reflection is different from inverse mapping. 
%Given two vectors $\vector{v}_m$ and $\vector{v}_w$, reflection can transfer both of them with identical reflection mapping as in Eqs. \ref{eq:inv1} and \ref{eq:inv2}, but not inverse mapping.
When $m$ and $w$ are paired words, reflection can transfer $\vector{v}_m$ and $\vector{v}_w$ each other with identical reflection mapping as in Eqs. \ref{eq:inv1} and \ref{eq:inv2}, but an inverse mapping cannot.
Given vector $\vector{v}$ in Euclidean space $\mathbb{R}^n$, the formula for the reflection in the mirror is given:
\begin{eqnarray}
    \mathrm{Ref}_{\vector{a},\vector{c}}(\vector{v}) = \vector{v} - 2 \frac{(\vector{v} - \vector{c}) \cdot \vector{a}}{\vector{a} \cdot \vector{a}}\vector{a},
	\label{eq:ref}
\end{eqnarray}
where $\vector{a} \in \mathbb{R}^n$ is a vector orthogonal to the mirror and $\vector{c} \in \mathbb{R}^n$ is a point through which the mirror passes. 
$\vector{a}$ and $\vector{c}$ are parameters that determine the mirror.

\subsection{Proposed method: Reflection-based Word Attribute Transfer}
%\subsubsection{Reflection by a Single Mirror}
\label{non_pm}
\begin{figure}[h]
   \centering
    \includegraphics[clip, width=5cm]{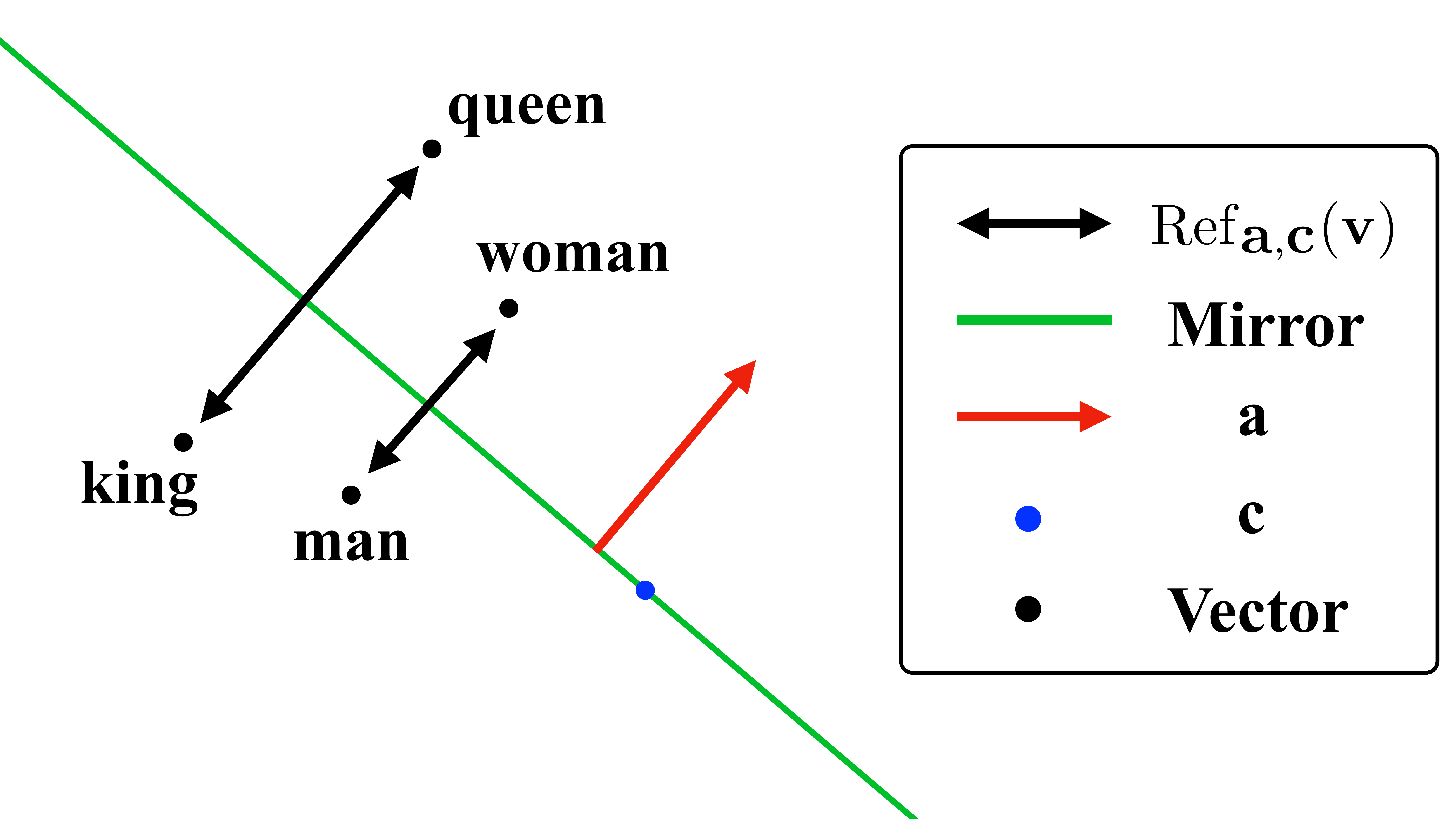}
    \caption{Reflection-based word attribute transfer with a single mirror}
    \label{fig:ref}
\end{figure}
We apply reflection to the word attribute transfer.
We learn a mirror (hyperplane) in a pre-trained embedding space using training word pairs with binary attribute $\vector{z}$ (Fig. \ref{fig:ref}).
Since the mirror is uniquely determined by two parameter vectors, $\vector{a}$ and $\vector{c}$, we estimate $\vector{a}$ and $\vector{c}$ from target attribute $\vector{z}$ using fully connected multi-layer perceptrons:
\begin{eqnarray}
    \vector{a} & = & \mathrm{MLP}_{\theta_1}(\vector{z}), \label{eq:a} \\
    \vector{c} & = & \mathrm{MLP}_{\theta_2}(\vector{z}), \label{eq:c}
\end{eqnarray}
where $\theta$ is a set of trainable parameters of $\mathrm{MLP}_{\theta}$.
Here, $\theta_1$ and $\theta_2$ are optimized for each attribute dataset.
Transferred vector $\vector{v}_y$ is obtained by inverting attribute $\vector{z}$ of $\vector{v}_x$ by reflection:
\begin{eqnarray}
    \vector{v}_y = \mathrm{Ref}_{\vector{a},\vector{c}}(\vector{v}_x).
    \label{eq:y}
\end{eqnarray}

%\subsubsection{Reflection by Parameterized Mirrors}
%
%\paragraph{}
\label{pm}
\begin{figure}[h]
   \centering
    \includegraphics[clip, width=5cm]{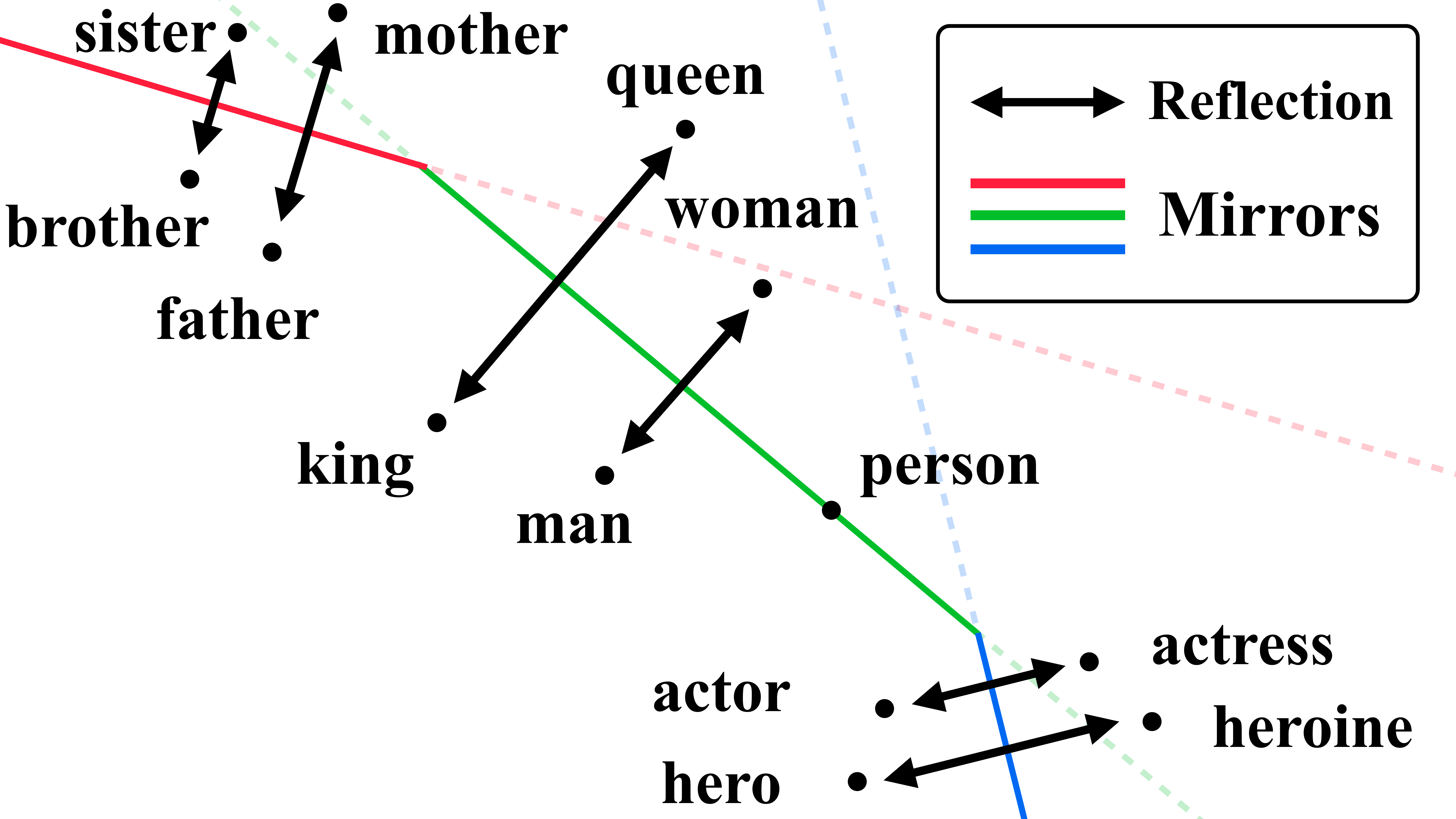}
    \caption{Reflection with parameterized mirrors} 
    \label{fig:pm}
\end{figure}
Reflection with a mirror by Eqs. \ref{eq:a} and \ref{eq:c} assumes a single mirror that only depends on $\vector{z}$.
%
%Previous discussion assumed that there will be pairs sharing a stable pair such as \emph{king} and \emph{queen}. 
Previous discussion assumed pairs that share a stable pair, such as \emph{king} and \emph{queen}. 
%
%However, gendered words often do not come in pairs, and the gender is not stable enough to be modeled by a single mirror. 
However, since gendered words often do not come in pairs, gender is not stable enough to be modeled by a single mirror.
For example, although \emph{actress} is exclusively feminine, \emph{actor} is clearly neutral in many cases.
Thus, \emph{actor} is not obviously a masculine counterpart like \emph{king}.
In fact, bias exists in gender words in the embedding space \cite{DBLP:conf/emnlp/ZhaoZLWC18, DBLP:conf/acl/KanekoB19}.
This phenomenon can occur not only with gender attributes but also with other attributes.
The single mirror assumption forces the mirror to be a hyperplane that goes through the midpoints for all the word vector pairs.
However, the vector pair \emph{actor}-\emph{actress} shown on the right in Fig. \ref{fig:pm} cannot be transferred well since the single mirror (the green line) does not satisfy this constraint due to the bias of the embedding space.  
To solve this problem, we propose \textit{parameterized mirrors}, based on the idea of using different mirrors for different words.
We define mirror parameters $\vector{a}$ and $\vector{c}$ using word vector $\vector{v}_x$ to be transferred in addition to attribute vector $\vector{z}$:
\begin{eqnarray}
    \vector{a} & = & \mathrm{MLP}_{\theta_1}([\vector{z} ; \vector{v}_x]) \label{eq:a_pm}, \\
    \vector{c} & = & \mathrm{MLP}_{\theta_2}([\vector{z} ; \vector{v}_x]) \label{eq:c_pm},
\end{eqnarray}
where $[\cdot ; \cdot]$ indicates the vector concatenation in the column. 
The \textit{parameterized mirrors} are expected to work more flexibly on different words than a single mirror because \textit{parameterized mirrors} dynamically determine similar mirrors for similar words.
For instance, as shown in Fig. \ref{fig:pm}, suppose we learned the mirror (the blue line) that transfers $\vector{v}_{hero}$ to $\vector{v}_{heroine}$ in advance.
If input word vector $\vector{v}_{actor}$ resembles $\vector{v}_{hero}$, a mirror that resembles the one for $\vector{v}_{hero}$ should be derived and used for the transfer.

On the other hand, the reflection works as an identity mapping for a vector on the mirror (e.g., $\vector{v}_{person}$ in Fig \ref{fig:pm}). That is, the proposed method assumes that non-attribute word vectors are located on the mirror. %%% sudoh; camera-ready
Since we used a 300-dimensional embedded space in the experiment, we assume that the non-attribute word vector exists in a 299-dimensional subspace.

Here, it should be noted that Eq. \ref{eq:doubleRef} may not hold for parameterized mirrors. 
In reflection with a single mirror, it is true that $\vector{v} = \mathrm{Ref}_{\vector{a},\vector{c}}(\hspace{0.7mm} \mathrm{Ref}_{\vector{a},\vector{c}}(\vector{v}))$.
However, with the $\vector{v}$-parameterized reflection $\mathrm{Ref}_{\vector{a_v},\vector{c_v}}(\vector{v})$,
%twice computation is not ensured to equal to $\vector{v}$. 
this is not guaranteed.
%
%Because mirror parameters $\vector{a_v}$ and $\vector{c_v}$ is parameterized for the input word.
Because mirror parameters $\vector{a_v}$ and $\vector{c_v}$ depend on an input word vector as Eqs. \ref{eq:a_pm} and \ref{eq:c_pm}.
%In reflection with parameterized mirrors, the twice computation ensured by loss.
Thus, we exclude this constraint and employ the constraints given by Eqs. \ref{eq:inv1}-\ref{eq:inv_non} for our loss function.
%%% Thus, we exclude this constraint and employ the constraints given by Eqs. 5-7 for our loss function. とか． %%% sudoh; camera-ready
%\subsubsection{Loss Function}
\paragraph{}
The following property must be satisfied in word attribute transfer: (1) words with attribute $\vector{z}$ are transferred and (2) words without it are not transferred.
Thus, loss $L(\theta_1, \theta_2)$ is defined:
\begin{align}
    L(\theta_1, \theta_2) = 
    \frac{1}{|\mathcal{A}|} \sum_{(x,t,\vector{z}) \in \mathcal{A} }( \vector{v}_y - \vector{v}_t)^2 \label{eq:loss_A} \\
    +  \frac{1}{|\mathcal{N}|}  \sum_{(x,\vector{z}) \in \mathcal{N} }(\vector{v}_y - \vector{v}_x)^2, \label{eq:loss_N}
    %\label{eq:loss}
\end{align}
where Eq. \ref{eq:loss_A} is a term that draws target word vector $\vector{v}_{t_i}$ closer to corresponding transferred vector $\vector{v}_{y_i}$ and Eq. \ref{eq:loss_N} is a term that prevents words without a target attribute from being moved by transfer function.
$\vector{v}_y$ is the output of a reflection (Eq. \ref{eq:y}).

\section{Experiment}
We evaluated the performance of the word attribute transfer using data with four different attributes. 
%
%\section{Experimental Setup}
We used 300-dimensional word2vec and GloVe as the pre-trained word embedding.
We used four different datasets of word pairs with four binary attributes: Male-Female, Singular-Plural, Capital-Country, and Antonym (Table \ref{tab:dataset}).
%
%\footnote[3]{Note that these gender word pairs are an assumption because they contain socially problematic words.}
%
These word pairs were collected from analogy test sets \cite{DBLP:journals/corr/abs-1301-3781, DBLP:conf/naacl/GladkovaDM16} and the Internet. 
Noun antonyms were taken from the literature \cite{DBLP:conf/eacl/NguyenWV17}.
For non-attribute dataset $\mathcal{N}$, we sampled words from the vocabulary of word embedding.
We sampled from 4 to 50 words for training and 1000 for the test ($|\mathcal{N}_{\mathrm{test}}| = 1000$).
\begin{table}[ht]
\centering
\caption{Statistics of binary attribute word pair datasets (in number of word pairs)}
\label{tab:dataset}
\scalebox{0.85}[0.85]{ 
\begin{tabular}{ccccc}
\toprule
\textbf{Dataset} $\mathcal{A}$   & \textbf{Train}    & \textbf{Val}     & \textbf{Test}    & \textbf{Total}     \\ \midrule
Male-Female (MF)        & 29 & 12 & 12 & 53  \\ 
Singular-Plural (SP)    & 90 & 25 & 25 & 140 \\ 
Capital-Country (CC)    & 59 & 25 & 25 & 109 \\ 
Antonym (AN)            & 1354 & 290 & 290 & 1934 \\ \bottomrule
\end{tabular}
}
\end{table}
\subsection{Evaluation Metrics} 
\label{eval}
We measured the accuracy and stability performances of the word attribute transfer.
%We measured the accuracy and stability performances of the word attribute transfer\footnote[2]{Code are available at: \url{https://github.com/yoichi1484/reflection_based_word_attribute_transfer}}.
%
The accuracy measures how many input words in $\mathcal{A}_{\mathrm{test}}$ were transferred correctly to the corresponding target words.
The stability score measures how many words in $\mathcal{N}_{\mathrm{test}}$ were not mapped to other words.
For example, in the Male-Female transfer, given \emph{man}, the transfer is regarded as correct if \emph{woman} is the closest word to the transferred vector; otherwise it is incorrect.
Given \emph{person}, the transfer is regarded as correct if \emph{person} is the closest word to the transferred vector; otherwise it is incorrect.
The accuracy and stability scores are calculated by the following formula:
\begin{equation}
\scalebox{0.9}{$
\delta (\vector{v}_y, t) = 
  \begin{cases}
    1  & \hspace{2mm} \mathrm{if} \hspace{3mm} \argmax_{k \in \mathcal{V}} (\mathrm{cos}(\vector{v}_y, \vector{v}_k)) = t \\
    0  & \hspace{2mm} \mathrm{otherwise},
  \end{cases}\\
$}
\end{equation}
\begin{eqnarray}
	\mathrm{Accuracy} = \frac{1}{|\mathcal{A}_{\mathrm{test}}|} \sum_{(x, t, \vector{z}) \in \mathcal{A}_{\mathrm{test}}} \delta (\vector{v}_{y}, t),\\
	\mathrm{Stability} =  \frac{1}{|\mathcal{N}_{\mathrm{test}}|} \sum_{(x, \vector{z}) \in \mathcal{N}_{\mathrm{test}}} \delta (\vector{v}_{y}, x),
\end{eqnarray}
where $\mathcal{V}$ is the vocabulary of the word embedding model and $\mathrm{cos}(\vector{v}_y, \vector{v}_k)$ is the cosine similarity measure, defined as: $\mathrm{cos}(\vector{v}_y, \vector{v}_k) = \frac{\vector{v}_y \cdot \vector{v}_k}{\| \vector{v}_y\| \|\vector{v}_k\|}$.

\subsection{Methods and Configurations} 
In our experiment, we compared our proposed method with the following baseline methods\footnote[2]{Our code and datasets are available at: \url{https://github.com/ahclab/reflection}}:
\begin{description}
  \item[\textsc{Ref}]
  Reflection-based word attribute transfer with a single mirror. We used a fully connected 2-layer MLP with 300 hidden units and ReLU  \cite{DBLP:journals/jmlr/GlorotBB11} to estimate $\vector{a}$ and $\vector{c}$.
  \item[\textsc{Ref}$+$\textsc{PM}]
  Reflection-based word attribute transfer with \textit{parameterized mirrors}. We used the same architecture of MLP as the \textsc{Ref}. 
  \item[\textsc{MLP}]
  Fully connected MLP with 300 hidden units and ReLU: $\vector{v}_y = \mathrm{MLP} ([\vector{v}_x; \vector{z}])$. The highest accuracy models in SGNS are a 2-layer MLP for Capital-Country and 3-layer MLP for the other datasets. The highest accuracy models in GloVe are a 2-layer MLP for Singular-Plural and 3-layer MLP for the other datasets.
  \item[\textsc{Diff}]\
  Analogy-based word attribute transfer with a difference vector: $\vector{d} = \vector{v}_m - \vector{v}_w$, where $m$ and $w$ are in the training data of $\mathcal{A}$. We chose $\vector{d}$ that achieved the best accuracy in the validation data of $\mathcal{A}$.
  We determined whether to add or subtract $\vector{d}$ to $\vector{v}_x$ based on the explicit knowledge (Eq. \ref{eq:an3}). 
  \textsc{Diff}$^+$ and \textsc{Diff}$^-$ transfer with a difference vector regardless of the explicit knowledge.  $^+$ and $^-$ add or subtract the difference vector to any input word vector. %%カメラレディ追加2
  \item[\textsc{MeanDiff}]
   Analogy-based word attribute transfer with a mean difference vector $\bar{\vector{d}}$: $\bar{\vector{d}} = \frac{1}{|\mathcal{A}_{\mathrm{train}}|} \sum_{(m_i, w_i, \vector{z}) \in \mathcal{A}_{\mathrm{train}}} (\vector{v}_{m_i} - \vector{v}_{w_i})$.
  We determined whether to add or subtract $\bar{\vector{d}}$ to $\vector{v}_x$ based on the explicit knowledge (Eq. \ref{eq:an3}).
\end{description} 

For proposed methods, we used the Adam optimizer \cite{DBLP:journals/corr/KingmaB14} with $\alpha = 10^{-4}$ for Male-Female, Singular-Plural and Capital-Country, and $\alpha = 15^{-3}$ for Antonym (the other hyperparameters were the same as the original one \cite{DBLP:journals/corr/KingmaB14}).
We did not use such regularization methods as dropout \cite{DBLP:journals/jmlr/SrivastavaHKSS14} or batch normalization \cite{DBLP:conf/icml/IoffeS15} because they did not show any improvement in our pilot test.
We implemented \textsc{Ref}, \textsc{Ref}$+$\textsc{PM} and MLP with Chainer \cite{DBLP:conf/kdd/TokuiOANOSSUVV19}, which is one of the best deep learning frameworks.

\begin{table*}[ht!]
\centering
\caption{Results in accuracy and stability scores: MF, SP, CC, and AN are datasets.}
\label{tab:score_table}
%\scalebox{0.57}[0.57]{ 
\scalebox{0.65}[0.65]{ 
\begin{tabular}{lcrrrrrrrrrrrrrrrrrrrr}
\toprule
\multirow{3}{*}{\textbf{Method}} & \multirow{3}{*}{\textbf{Knowledge}} & \multicolumn{9}{c}{\textbf{word2vec}}                                                                                                     & & \multicolumn{9}{c}{\textbf{GloVe}}                                                                                                       \\ \cmidrule(r){3-11} \cmidrule(r){13-21}
                                 &                                     & \multicolumn{4}{c}{\textbf{Accuracy (\%)}}                        & & \multicolumn{4}{c}{\textbf{Stability (\%)}}                         & & \multicolumn{4}{c}{\textbf{Accuracy (\%)}}                        & & \multicolumn{4}{c}{\textbf{Stability (\%)}}                        \\ %\cmidrule(r){3-6} \cmidrule(r){8-11} \cmidrule(r){13-16} \cmidrule(r){18-21}
                                 &                                     & MF            & SP            & CC            & AN            & & MF            & SP             & CC            & AN              & & MF            & SP            & CC            & AN            & & MF             & SP             & CC             & AN              \\ \midrule     
\textsc{Ref}                     &                                     & 20.8          & 0.0           & 36.0          & 0.0           & & 99.8          & \textbf{100.0} & \textbf{99.8} & \textbf{100.0}  & & 12.5          & 2.0           & 26.0          & 0.0           & & \textbf{100.0} & \textbf{100.0} & \textbf{100.0} & \textbf{100.0}  \\
\textsc{Ref}$+$\textsc{PM}       &                                     & \textbf{41.7} & \textbf{22.0} & \textbf{58.0} & 28.8          & & \textbf{99.9} & 99.4           & 99.4          & \textbf{100.0}  & & \textbf{45.8} & \textbf{50.0} & \textbf{76.0} & 33.5          & & 99.7           & 99.1           & 99.2           & \textbf{100.0}  \\
MLP                              &                                     & 8.3           & 4.0           & 12.0          & \textbf{35.9} & & 2.2           & 0.0            & 2.7           & 1.9             & & 4.2           & 10.0          & 18.0          & \textbf{36.7} & & 5.1            & 7.0            & 5.2            & 1.2             \\
\textsc{Diff} $^+$               &                                     & 25.0          & 2.0           & 32.0          & -             & & 72.1          & 77.9           & 53.9          & -               & & 25.0          & 2.0           & 26.0          & -             & & 99.3           & 94.2           & 99.3           & -               \\
\textsc{Diff} $^-$               &                                     & 25.0          & 2.0           & 30.0          & -             & & 49.6          & 78.2           & 56.3          & -               & & 25.0          & 2.0           & 24.0          & -             & & \textbf{100.0}          & 99.9           & 99.5           & -               \\
\textsc{MeanDiff} $^+$           &                                     & 4.2           & 0.0           & 22.0          & -             & & 98.6          & 99.4           & 87.6          & -               & & 0.0           & 0.0           & 22.0          & -             & & \textbf{100.0} & \textbf{100.0} & \textbf{100.0} & -               \\
\textsc{MeanDiff} $^-$           &                                     & 8.3           & 0.0           & 14.0          & -             & & 97.2          & 99.3           & 92.4          & -               & & 0.0           & 0.0           & 0.0           & -             & & \textbf{100.0} & \textbf{100.0} & \textbf{100.0} & -               \\ \midrule
\textsc{Diff}                    & \checkmark                          & 62.5          & 4.0           & 64.0          & -             & & -             & -              & -             & -               & & 50.0          & 4.0           & 44.0          & -             & & -              & -              & -              & -               \\
\textsc{MeanDiff}                & \checkmark                          & 12.5          & 0.0           & 36.0          & -             & & -             & -              & -             & -               & & 0.0           & 0.0           & 0.0           & -             & & -              & -              & -              & -               \\ \bottomrule
\end{tabular}
}
\end{table*} 
\subsection{Evaluation in Accuracy and Stability}
Table \ref{tab:score_table} shows the accuracy and stability results. 
Different pre-trained word embeddings GloVe or word2vec gave similar results.
\textsc{Ref}$+$\textsc{PM} achieved the best accuracy among the methods that did not use explicit attribute knowledge.
For example, the accuracy of \textsc{Ref}$+$\textsc{PM} was 76\% in Capital-Country, but the accuracy of \textsc{Diff}$^+$ was 26\%.
For stability, reflection-based transfers achieved outstanding stability scores that exceeded 99\%.
%
%The stability of \textsc{Diff} $^+$ and \textsc{Diff} $^-$ was much lower than the other methods.
%
%Although \textsc{MeanDiff} $^-$ and \textsc{MeanDiff} $^+$ achieved high stability, their accuracy results were very low.
%
%Interestingly, reflection-based transfer with \textit{parameterized mirrors} (\textsc{Ref}$+$\textsc{PM}) achieved high performance in both accuracy and stability.
%
%For example, the accuracy of \textsc{Ref}$+$\textsc{PM} was 41.67\%, and the stability was 99.9\% in Male-Female (MF), and the accuracy was 58 \% and the stability was 99.40 \% in Capital-Country (CC). 
%
%These results show that the proposed method transfers an input word if it has a target attribute and does not transfer an input word, even though it does not use explicit attribute knowledge on the input words. 
%
The results show that our proposed method transfers an input word if it has a target attribute and does not transfer an input word with better score than the baselines otherwise, even though the proposed method does not use attribute knowledge of the input words. 
MLP worked poorly both in accuracy and stability. 
On the antonym dataset, although the transfer accuracy by the proposed method was a bit lower than that by MLP, the proposed method’s stability was 100\% and that of MLP was extremely poor: about 1\%.

\begin{figure}[h!]
\centering
    \includegraphics[clip, width=7cm]{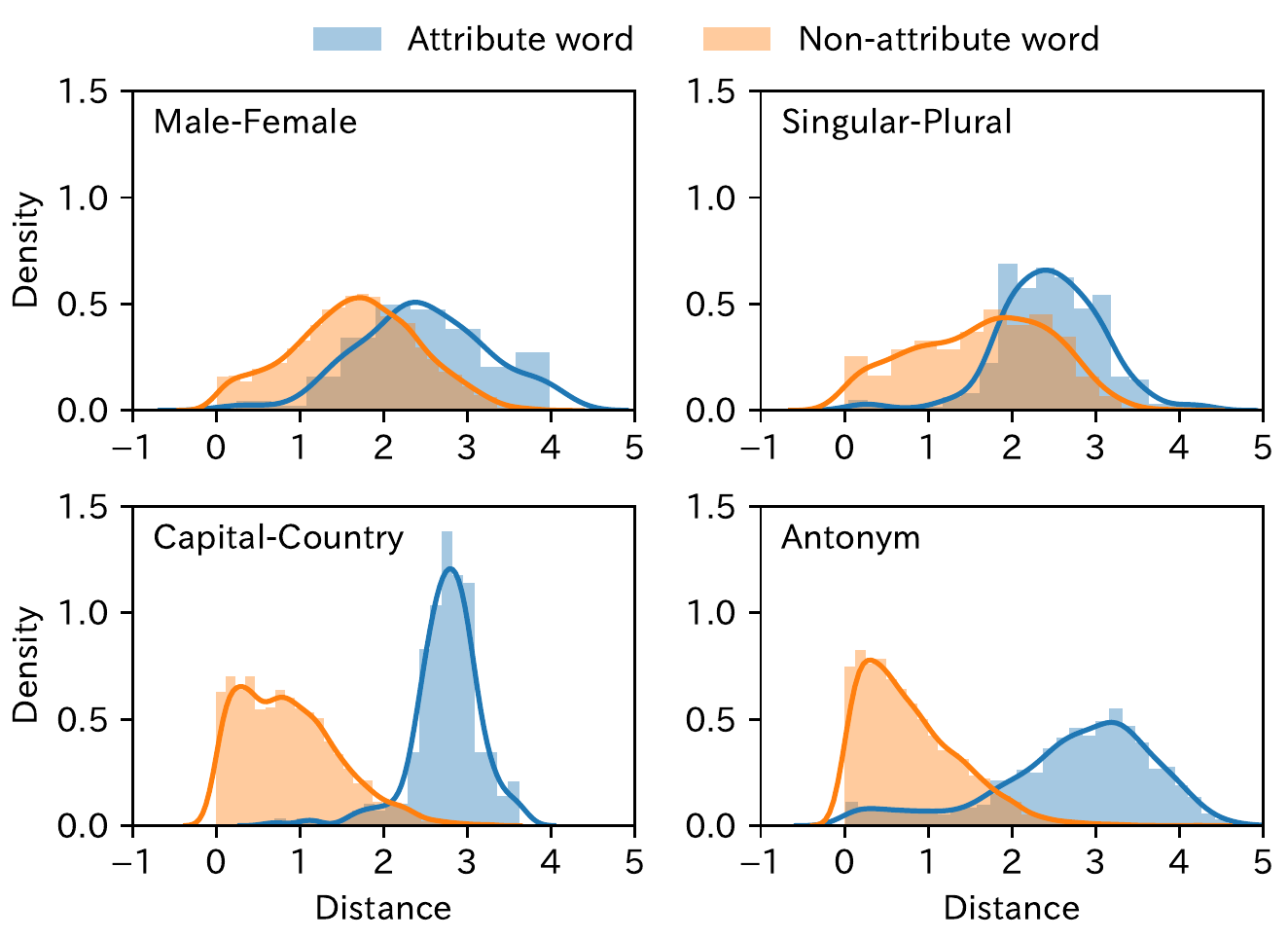}
    \caption{Distribution of distance between input word vector and its mirror  $\frac{|(\vector{v}_x-\vector{c}) \cdot \vector{a}|}{\| \vector{a} \|}$ learned by \textsc{Ref}$+$\textsc{PM}. Non-attribute words are close to the mirror, and attribute words are distributed away from it.} 
    \label{fig:distance}
\end{figure}

\begin{figure}[h!]
\centering
    \includegraphics[clip, width=7cm]{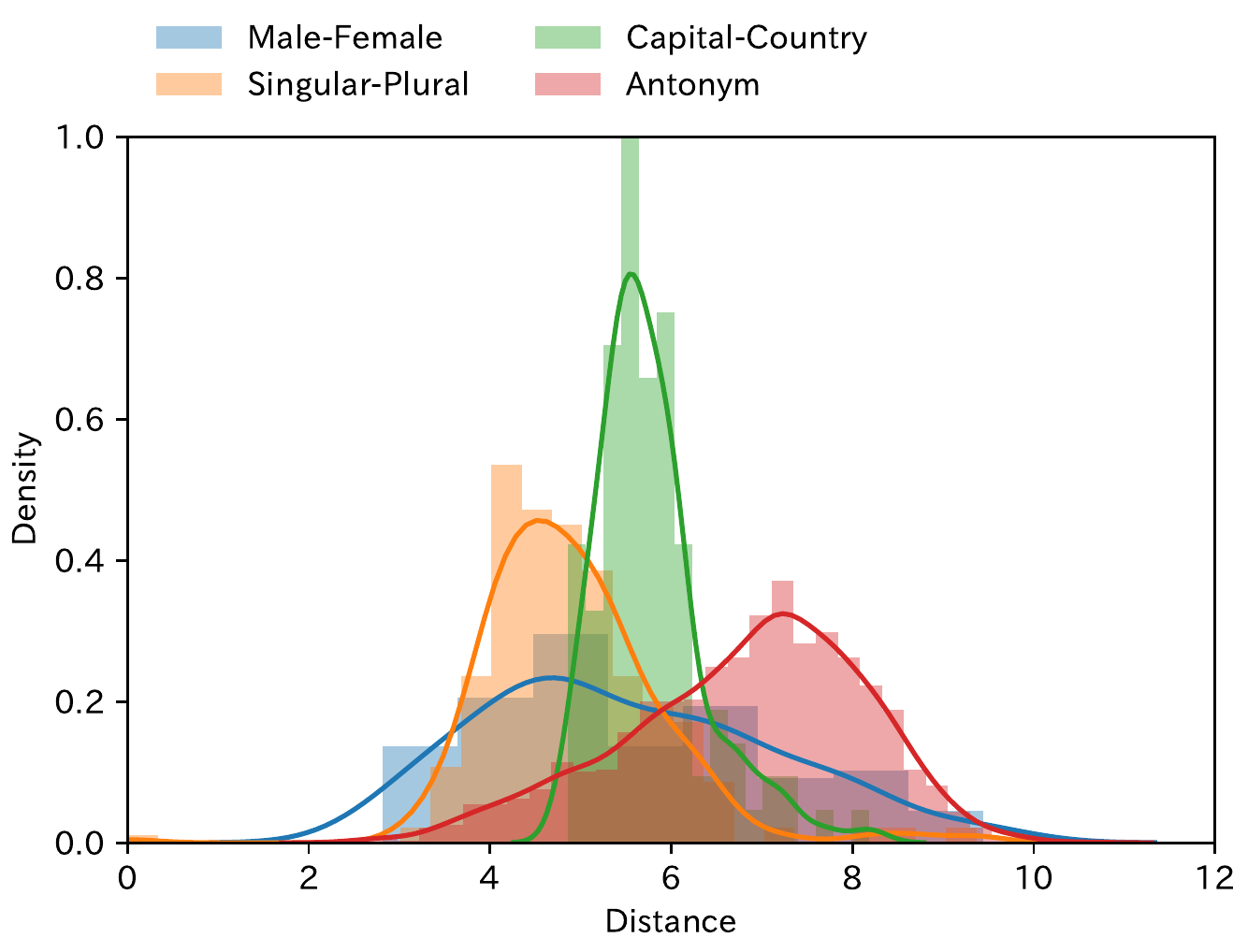}
    \caption{Distribution of distance between input word vector  $\vector{v}_x$ and target word vector $\vector{v}_t$}
    \label{fig:distance_w2w}
\end{figure}

\begin{table}[htbp]
\centering
\caption{Relation among size of $|\mathcal{N}_{\mathrm{train}}|$ and stability of learning-based methods}
\label{tab:stb}
\scalebox{0.6}[0.6]{ 
\begin{tabular}{clrrrrrrrrr}
\toprule
                             &                               & \multicolumn{4}{c}{\textbf{Accuracy (\%)}}                    & & \multicolumn{4}{c}{\textbf{Stability (\%)}}                       \\ \cmidrule(r){3-6} \cmidrule(r){8-11}
                             &                               & \multicolumn{4}{c}{$|\mathcal{N}_{\mathrm{train}}|$}          & & \multicolumn{4}{c}{$|\mathcal{N}_{\mathrm{train}}|$}              \\      
                             &                               & 0             & 4             & 10            & 50            & &  0             & 4              & 10             & 50             \\ \midrule
\multirow{3}{*}{\textbf{MF}} & \textsc{Ref}                  & 12.5          & 12.5          & 12.5          & 12.5          & & \textbf{100.0} & \textbf{100.0} & \textbf{100.0} & \textbf{100.0} \\
                             & \textsc{Ref}$+$\textsc{PM}    & \textbf{45.8} & \textbf{41.7} & \textbf{37.5} & \textbf{41.7} & & 99.7           & 99.9           & 99.9           & 99.9           \\
                             & MLP                           & 0.0           & 4.2           & 0.0           & 4.2           & & 0.0            & 0.4            & 1.0            & 5.0            \\ \midrule
                             
\multirow{3}{*}{\textbf{SP}} & \textsc{Ref}                  & 0.0           & 0.0           & 2.0           & 0.0           & & \textbf{100.0} & \textbf{100.0} & \textbf{100.0} & \textbf{100.0} \\
                             & \textsc{Ref}$+$\textsc{PM}    & \textbf{48.0} & \textbf{40.0} & \textbf{50.0} & \textbf{46.0} & & 53.3           & 99.1           & 99.1           & 99.8           \\
                             & MLP                           & 4.0           & 6.0           & 6.0           & 10.0          & & 0.0            & 0.5            & 1.7            & 7.0            \\ \midrule

\multirow{3}{*}{\textbf{CC}} & \textsc{Ref}                  & 24.0          & 26.0          & 24.0          & 20.0          & & \textbf{100.0} & \textbf{100.0} & \textbf{100.0} & \textbf{100.0} \\
                             & \textsc{Ref}$+$\textsc{PM}    & \textbf{76.0} & \textbf{72.0} & \textbf{74.0} & \textbf{74.0} & & 99.2           & \textbf{100.0} & 99.9           & 99.9           \\
                             & MLP                           & 16.0          & 10.0          & 14.0          & 18.0          & & 0.0            & 0.4            & 1.0            & 5.2            \\ \midrule

\multirow{3}{*}{\textbf{AN}} & \textsc{Ref}                  & 0.0           & 0.0           & 0.0           & 0.0           & & \textbf{100.0} & \textbf{100.0} & \textbf{100.0} & \textbf{100.0} \\
                             & \textsc{Ref}$+$\textsc{PM}    & 26.9          & 26.7          & 33.5          & 25.7          & & \textbf{100.0} & \textbf{100.0} & \textbf{100.0} & \textbf{100.0} \\
                             & MLP                           & \textbf{29.5} & \textbf{29.7} & \textbf{36.7} & \textbf{36.6} & & 0.1            & 0.5            & 1.2            & 4.6            \\ \bottomrule
\end{tabular}
}
\end{table} 
We investigated the relation between the training data size of the non-attribute words, and the stability of the learning-based methods by conducting an additional experiment that varied $|\mathcal{N}_{\mathrm{train}}|$. 
The stability scores by MLP did not improve (Table \ref{tab:stb}).
On the other hand, \textsc{Ref}$+$\textsc{PM} achieved high stability scores with $|\mathcal{N}_{\mathrm{train}}|=0$ and maintained the accuracy.
We hypothesized that the high stability came from the distance between the word and its mirror.
If non-attribute words are distributed on the mirror, they will not be transferred.
We investigated the distance between input word vector $\vector{v}_x$ and its mirror (Fig.~\ref{fig:distance}).
The result shows that non-attribute words are close to the mirror, and attribute words are distributed away from it.
In Male-Female and Singular-Plural, the distance is not significantly farther than Antonym and Capital-Country.
If the distance between paired words is very small,
the distance between the word and its mirror is also small.
Fig.~\ref{fig:distance_w2w} shows the distribution of the distance between input $\vector{v}_x$ and target word vector $\vector{v}_t$.
The distance of Male-Female and Singular-Plural is much smaller than Capital-Country and Antonym.

\subsection{Visualization of Parameterized Mirrors}
\begin{figure}[htbp!]
\centering
    \includegraphics[clip, width=6.8cm]{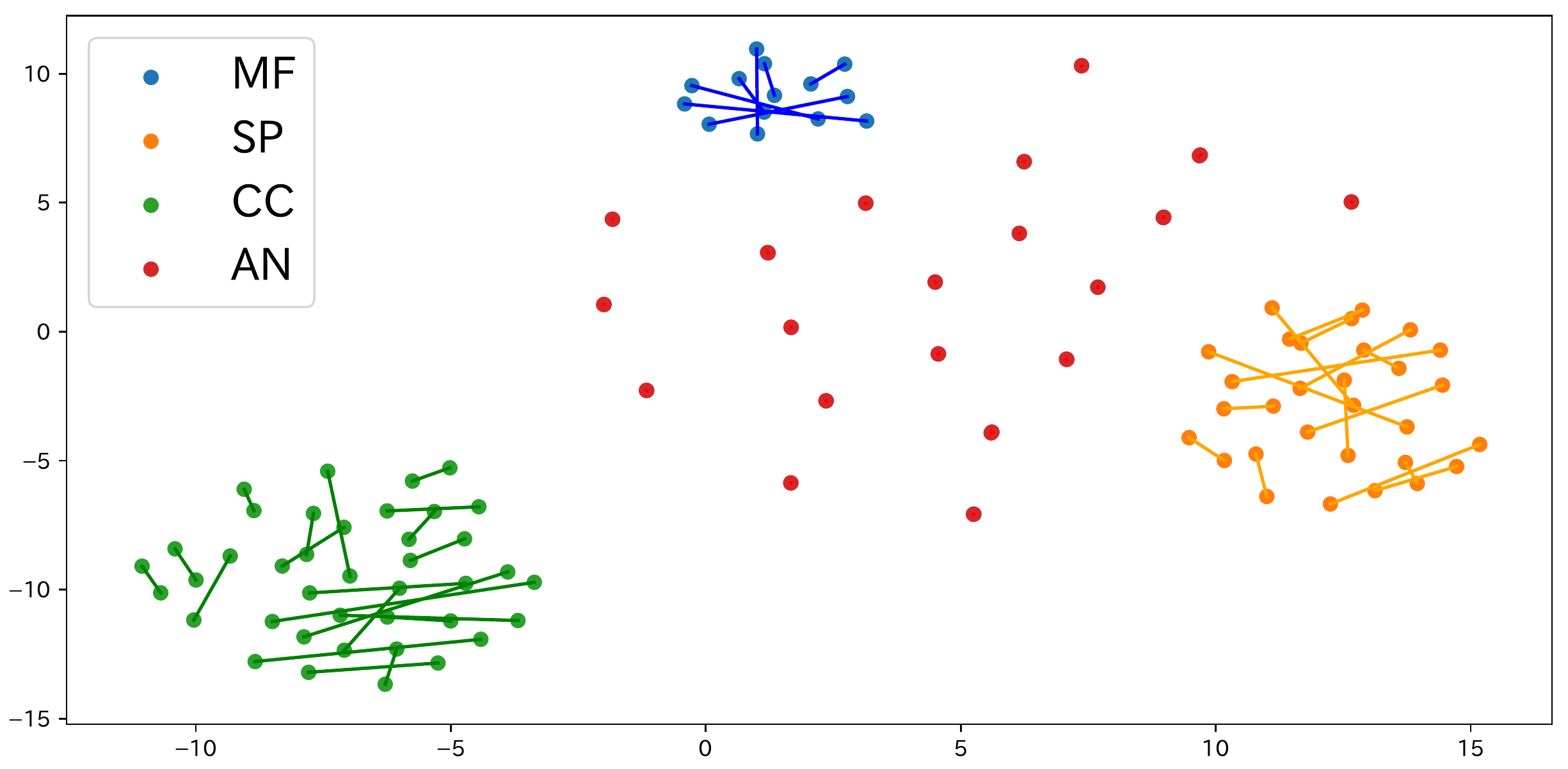}
    \caption{Two-dimensional t-SNE projection of $\vector{a}$}
    \label{fig:GloVe_a}
\end{figure}

Figure~\ref{fig:GloVe_a} shows the t-SNE results of mirror parameter $\vector{a}$ obtained for the test words.
Paired mirror, $(\vector{a}_x, \vector{a}_t)$, is connected by a line segment.
%
%Fig.~\ref{fig:GloVe_a} suggests not only the mirror parameters of paired words are similar to each other but also the parameters with the attribute form a cluster — words with the same attribute has similar mirror parameter $\vector{a}$.
Fig.~\ref{fig:GloVe_a} suggests that the mirror parameters of the paired words are similar to each other and that those with the attribute form a cluster; words with the same attribute have similar mirror parameters $\vector{a}$.

\subsection{Transfer Example}
Table~\ref{tab:example1} shows the gender transfer results for a tiny example sentence. 
Here the attribute transfer was applied to every word in the sentence.
MLP made many wrong transfers. % on words without gender attributes.
Analogy-based transfers can transfer only in one direction.
\textsc{Ref}$+$\textsc{PM} can transfer only attribute words. 
Table \ref{tab:example_ma} shows that words with different target attributes were transferred by each reflection-based transfer.
%
%Given \emph{actress} for a Male-Female transfer, it was transferred to \emph{actor} and to \emph{actors} for Singular-Plural.
%Given \emph{Tokyo} for Male-Female, Singular-Plural, and Antonym, it was not transferred, but it was transferred to \emph{Japan} for Country-Capital.
%Given \emph{rich} for Male-Female, Singular-Plural, and Capital-Country, it was not transferred, but it was transferred to \emph{poor} for Antonym.
%
%
\begin{table}[htbp!]
\centering
%\caption{Comparison of gender transfers. } 
\caption{Comparison of gender transfers. Each method transfers words in a sentence one by one. } 
\label{tab:example1}
\scalebox{0.68}[0.68]{ 
\begin{tabular}{l|l}
\toprule
$X$ & 
\begin{tabular}{l}
    the woman got married when you were a boy.
\end{tabular} \\ \midrule

\textsc{Ref} & 
\begin{tabular}{l}
    the \textbf{woman} got married when you were a \textbf{boy}.
\end{tabular} \rule[0mm]{0mm}{4mm} \\

\textsc{Ref}$+$\textsc{PM} & 
\begin{tabular}{l}
    the \textbf{man} got married when you were a \textbf{girl}.
\end{tabular} \rule[0mm]{0mm}{4mm} \\

%\begin{tabular}{l}
%\end{tabular} \\ 
%
%applied once &
%\begin{tabular}{l}
%    the \textbf{man} got married when you were a \textbf{girl}.
%\end{tabular} \rule[0mm]{0mm}{0mm} \\
%
%applied twice &
%\begin{tabular}{l}
%    the \textbf{woman} got married when you were a \textbf{boy}.
%\end{tabular} \rule[0mm]{0mm}{2mm} \\

\textsc{Diff} $^+$ & 
\begin{tabular}{l}
    the \textbf{man} got married when you were a \textbf{boy}.
\end{tabular} \rule[0mm]{0mm}{4mm} \\

\textsc{Diff} $^-$& 
\begin{tabular}{l} 
    she \textbf{woman} got married she you were a \textbf{girl}.
\end{tabular} \rule[0mm]{0mm}{4mm} \\ 

\textsc{MLP} & 
\begin{tabular}{l}
   By\_Katie\_Klingsporn \textbf{girlfriend} Valerie\_Glodowski\\
   fiancee Doughty\_Evening\_Chronicle ma'am \\
   Bob\_Grossweiner\_\& a \textbf{mother}.
\end{tabular} \rule[5mm]{0mm}{4mm} \\ \bottomrule
\end{tabular}
}
\end{table} 
\begin{table}[htbp!]
\centering
\caption{Transfer of different attributes with \textsc{Ref}$+$\textsc{PM}}
\label{tab:example_ma}
\scalebox{0.65}[0.65]{ 
\begin{tabular}{l|l}
\toprule
$X$ & 
\begin{tabular}{l}
    the rich actor wants to visit the beautiful city in tokyo.
\end{tabular} \\ \midrule
$+$ MF     & 
\begin{tabular}{l}
    the rich \textbf{actress} wants to visit the beautiful city in tokyo.
\end{tabular}\rule[0mm]{0mm}{4mm} \\ 
$+$ SP & 
\begin{tabular}{l}
    the rich \textbf{actresses} wants to visit the beautiful \textbf{cities} in tokyo.
\end{tabular} \rule[0mm]{0mm}{4mm} \\ 
$+$ CC & 
\begin{tabular}{l}
    the rich actresses wants to visit the beautiful cities in \textbf{japan}.
\end{tabular} \rule[0mm]{0mm}{4mm} \\ 
$+$ AN & 
\begin{tabular}{l} 
   the \textbf{poor}  actresses wants to visit the beautiful cities in japan.
\end{tabular} \rule[0mm]{0mm}{4mm} \\ \bottomrule
\end{tabular}
}
\end{table}

\section{Related Work}
%The embedded vectors obtained by SGNS \citep{DBLP:journals/corr/abs-1301-3781, DBLP:conf/nips/MikolovSCCD13} and GloVe \citep{DBLP:conf/emnlp/PenningtonSM14} have analogic relations.
%
The theory of analogic relations in word embeddings has been widely discussed \citep{DBLP:conf/nips/LevyG14, DBLP:journals/tacl/AroraLLMR16, DBLP:conf/acl/GittensAM17, DBLP:conf/acl/EthayarajhDH19a, DBLP:conf/icml/AllenH19,DBLP:conf/repeval/Linzen16}.
%
%\citet{DBLP:conf/nips/LevyG14} offer the explanation that SGNS factorizes a shifted PMI matrix.
%
%\citet{DBLP:conf/icml/AllenH19} and \citet{DBLP:conf/acl/EthayarajhDH19a} argued that they proved the existence of such analogic relations without strong assumptions. 
%
In our work, we focus on the analogic relations in a word embedding space and propose a novel framework to obtain a word vector with inverted attributes.
The style transfer task \citep{DBLP:conf/coling/NiuRC18, DBLP:conf/acl/TsvetkovBSP18, DBLP:conf/nips/LogeswaranLB18, DBLP:conf/aaai/JainMAS19, DBLP:conf/acl/DaiLQH19, DBLP:conf/iclr/LampleSSDRB19} resembles ours.
In style transfer, the text style of the input sentences is changed.
For instance, \citet{DBLP:conf/aaai/JainMAS19} transferred from formal to informal sentences. %formal sentence to informal sentence.
These style transfer tasks use sentence pairs;
our word attribute transfer task uses word pairs.
Style transfer changes sentence styles, but our task changes the word attributes. %
\citet{DBLP:conf/naacl/SoricutO15} studied morphological transformation based on character information. 
Our work aims for more general attribute transfer, such as gender transfer and antonym, and is not limited to morphological transformation.

\section{Conclusion}
%
%We proposed a reflection-based method for word attribute transfers without relying on the explicit attribute knowledge of an input word, which is necessary for a simple analogy-based transfer. 
%
%We proposed a word attribute transfer that does not rely on the explicit knowledge of input words. 
%
%
%In this paper, we proposed a novel method of word attribute transfer. %based on an ideal mapping that is called reflection.
%
This research aims to transfer word binary attributes (e.g., gender) for applications such as data augmentation of a sentence.
%
%In word analogy-based method, we need knowledge whether the word has the attribute or not (e.g., \emph{man} $\in$ gender, \emph{woman} $\in$ gender, \emph{person} $\notin$ gender).
We can transfer the word attribute with analogy of word vectors, but it
requires explicit knowledge whether the input word has the attribute or not (e.g., \emph{man} $\in$ gender, \emph{woman} $\in$ gender, \emph{person} $\notin$ gender).
%
%%% without attribute knowledge in inference time. %%% sudoh; camera-ready
The proposed method transfers binary word attributes using reflection-based mappings and keeps non-attribute words unchanged,
without attribute knowledge in inference time.
The experimental results showed that the proposed method outperforms analogy-based and MLP baselines in transfer accuracy for attribute words and stability for non-attribute words. % accuracy of attribute word transfer は accuracy of word attribute transfer ?

\bibliography{acl2020}
\bibliographystyle{acl_natbib}

\end{document}